\documentclass[letterpaper]{article} 
\usepackage{aaai23}  
\usepackage{times}  
\usepackage{helvet}  
\usepackage{courier}  
\usepackage[hyphens]{url}  
\usepackage{graphicx} 
\usepackage{natbib}  
\usepackage{caption} 
\usepackage[ruled,linesnumbered]{algorithm2e}
\usepackage{amsmath}
\usepackage{mathtools}
\usepackage{comment}
\usepackage{amsfonts}
\usepackage{amssymb}
\usepackage{multirow}
\usepackage{xcolor}

\newcommand{\methodname}{\textsc{FedOBD}}

\newcommand{\vect}[1]{\boldsymbol{#1}}
\newcommand{\norm}[1]{\left\lVert#1\right\rVert}
\newcommand{\funname}[1]{\texttt{#1}}

\newcommand{\lonenorm}[1]{\lvert{} #1 \rvert}

\title{Efficient Training of Large-scale Industrial Fault Diagnostic Models through Federated  Opportunistic Block Dropout}

\author {
  Yuanyuan Chen\textsuperscript{\rm 1}\equalcontrib,
  Zichen Chen\textsuperscript{\rm 1,2}\equalcontrib,
  Sheng Guo\textsuperscript{\rm 3}\equalcontrib,
  Yansong Zhao\textsuperscript{\rm 1},
  Zelei Liu\textsuperscript{\rm 1},
  Pengcheng Wu\textsuperscript{\rm 1},
  Chengyi Yang\textsuperscript{\rm 3},
  Zengxiang Li\textsuperscript{\rm 3$\dagger$},
  Han Yu\textsuperscript{\rm 1$\dagger$}
}
\affiliations {
  \textsuperscript{\rm 1} School of Computer Science and Engineering, Nanyang Technological University, Singapore\\
  \textsuperscript{\rm 2} University of California, Santa Barbara, CA, USA\\
  \textsuperscript{\rm 3} Digital Research Institute, ENN Group, Beijing, China\\
  \textsuperscript{$\dagger$}Corresponding authors: 	lizengxiang@enn.cn, han.yu@ntu.edu.sg
}

\usepackage{bibentry}

\begin{document}

\maketitle

\begin{abstract}
  Artificial intelligence (AI)-empowered industrial fault diagnostics is important in ensuring the safe operation of industrial applications. Since complex industrial systems often involve multiple industrial plants (possibly belonging to different companies or subsidiaries) with sensitive data collected and stored in a distributed manner, collaborative fault diagnostic model training often needs to leverage federated learning (FL). As the scale of the industrial fault diagnostic models are often large and communication channels in such systems are often not exclusively used for FL model training, existing deployed FL model training frameworks cannot train such models efficiently across multiple institutions. In this paper, we report our experience developing and deploying the \underline{Fed}erated \underline{O}pportunistic \underline{B}lock \underline{D}ropout (\methodname) approach for industrial fault diagnostic model training. By decomposing large-scale models into semantic blocks and enabling FL participants to opportunistically upload selected important blocks in a quantized manner, it significantly reduces the communication overhead while maintaining model performance.
  Since its deployment in ENN Group in February 2022, \methodname{} has served two coal chemical plants across two cities in China to build industrial fault prediction models. It helped the company reduce the training communication overhead by over 70\% compared to its previous AI Engine, while maintaining model performance at over 85\% test F1 score. To our knowledge, it is the first successfully deployed dropout-based FL approach.
\end{abstract}

\section{Introduction}
In modern industries, machinery is becoming increasingly sophisticated and facing highly demanding operational conditions. For example, rotating machinery (e.g., turbines, fans and pumps) are key components that are widely used in power generation and chemical plants. Slight performance deterioration, if not addressed early, could lead to sudden breakdowns or even serious accidents involving significant financial losses and/or human casualty.
As industries modernize towards the vision of Industry 4.0 \cite{Ghobakhloo:2020}, a wide variety of sensing devices are starting to be deployed in industrial settings to help monitor important equipment. The data collected by such devices make it possible to train intelligent fault diagnostic models for system maintenance decision support.

Artificial intelligence (AI) technologies have increasingly been applied in industrial fault diagnostic model training. The performance of such machine learning-based solutions depends on having access to large amounts of high quality data. However, data from a single factory might not be adequate to train such models effectively. As data are often collected and owned by different organizations in a given field, collaborative model training \cite{Warnat-Herresthal-et-al:2021} has been recognized as a useful technique to improve the quality of AI solutions in such situations.

As societies become increasingly aware of data privacy protection issues (e.g., following the introduction of data privacy regulations such as the General Data Protection Regulation (GDPR) \cite{GDPR}), federated learning (FL) \cite{FL2019,kairouz2019advances} - a privacy-preserving collaborative machine learning paradigm - has emerged. It has been rapidly gaining traction and has been applied in wide-ranging applications including safety management \cite{liu2020fedvision}, banking \cite{Long-et-al:2020} and smart healthcare \cite{Liu-et-al2022IAAI}.

In recent years, industrial fault detection applications powered by FL are starting to emerge \cite{Ma-et-al:2021,Zhang-et-al:2021,Geng-et-al:2022,Wang-et-al:2022}. These applications generally build on top of the popular Federated Averaging (FedAvg) FL model aggregation approach \cite{pmlr-v54-mcmahan17a}. Although they are useful for supporting privacy-preserving collaborative model training, they are not optimized for training large-scale deep neural network (DNN) models which are commonly required to build effective industrial fault diagnostic models \cite{Liu-et-al:2022IJAMT}. This is exacerbated by constraints on bandwidth usage for FL model training imposed by industries as the communication channel is often shared by multiple applications, some of which are safety critical.

To address the aforementioned challenges facing FL-based industrial fault diagnostics solutions, we propose the \textit{\underline{Fed}erated \underline{O}pportunistic \underline{B}lock \underline{D}ropout (\methodname)} approach \cite{FedOBD}. Its advantages are as follows:
\begin{enumerate}
  \item \textbf{Training Large-scale DNNs Efficiently}: \methodname{} divides a DNN into semantic blocks. Based on evaluating the importance of the blocks (instead of determining individual parameter importance like in the cases of~\cite{fed-dropout,FedDropoutAvg}), it opportunistically discards unimportant blocks in order to drastically reduce the size of the resulting model. Combined with parameter quantization, \methodname{} can significantly reduce the communication overhead incurred during FL model training.
  \item \textbf{Preserving Model Performance}: As blocks that are most important to the performance of the FL model are retained, \methodname{} can preserve model performance.
  \item \textbf{Supporting Incentive Distribution}: By storing the historical records of the important blocks contributed to FL model training by each data owner, \methodname{} can provide useful information for contribution-based incentive allocation \cite{VCG2020,Yu-et-al:2020AIES}.
\end{enumerate}
Compared to existing efficient FL model training frameworks, \methodname{} offers new capabilities which can support more sophisticated use cases.

The \methodname{} approach has been deployed through a collaboration between \textit{ENN Group}\footnote{\url{https://www.enn.cn/}} and the Trustworthy Federated Ubiquitous Learning (TrustFUL) Lab\footnote{\url{https://trustful.federated-learning.org/}}, Nanyang Technological University (NTU), Singapore since February 2022. It is used to replace the FedAvg model aggregation approach in the ENN FL model training platform to support FL model training under server-based horizontal FL settings \cite{FL2019}. In such scenarios, FL participants' datasets have large overlaps in the feature space, but little overlap in the sample space.
It has helped ENN Group train intelligent industrial fault diagnostic models involving two factories from two cities in China.
Throughout the deployment period, \methodname{} has helped ENN Group reduce the training communication overhead by over 70\% compared to its previous implementation, while maintaining model performance at over 85\% test F1 score. To the best of our knowledge, it is the first dropout-based federated learning method successfully deployed in industrial settings.

With the help of \methodname{}, the ENN Group federated learning subsystem has avoided the problems of long delays of training/updating industrial fault prediction models through FL, while maintaining comparable model performance. This capability allows it to provide more rapid update of such models to its enterprise customers and subsidiaries, thereby improving safe operation.

\section{Application Description}
\begin{figure}[t!]
  \centering
  \includegraphics[clip, trim=0cm 9.5cm 21cm 0cm, width=1\linewidth]{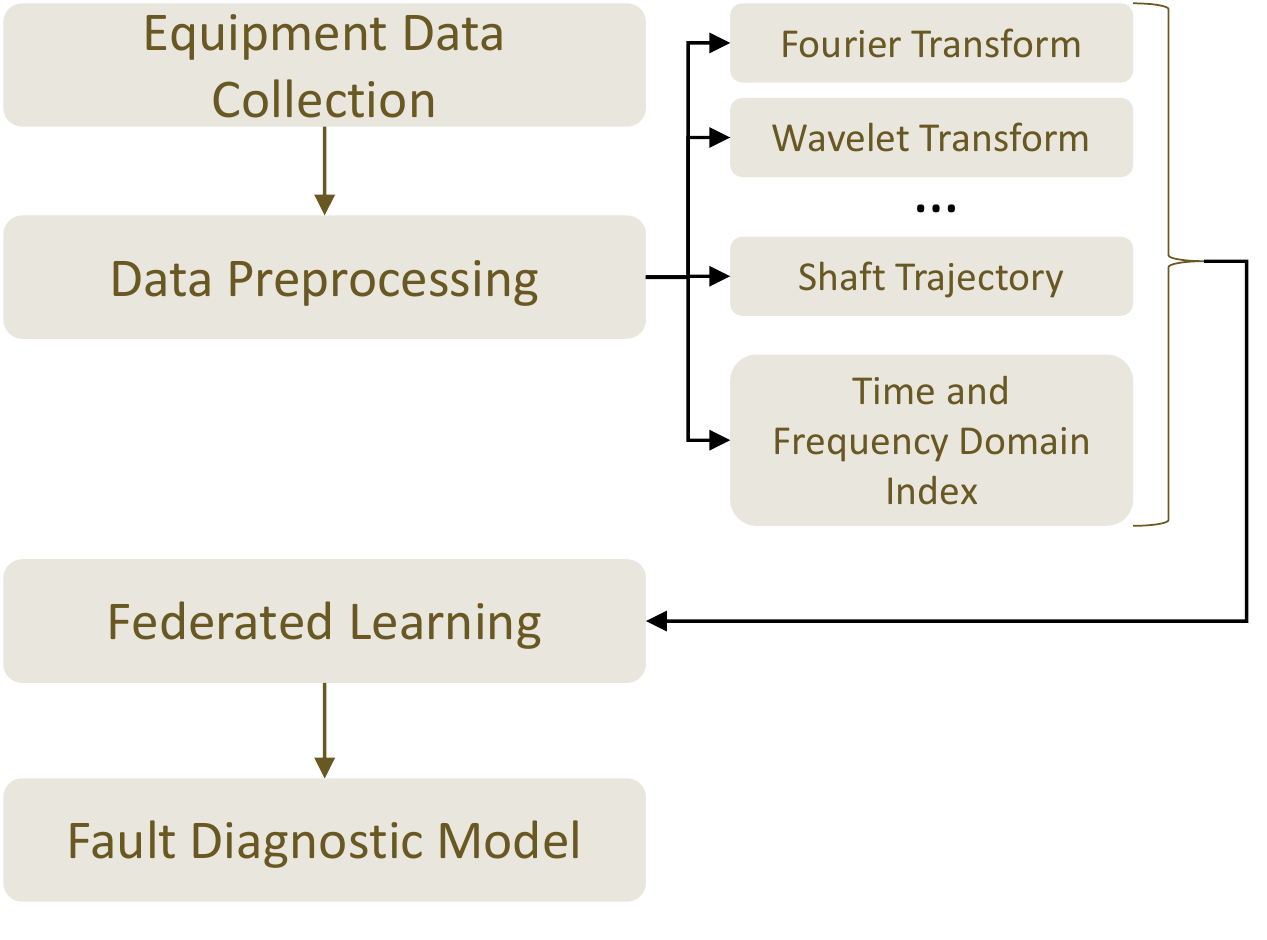}
  \caption{An overview of the ENN intelligent industrial fault diagnostics platform.} \label{fig:Overview}
\end{figure}

\begin{figure*}[t!]
  \centering
  \includegraphics[clip, trim=0cm 10cm 0cm 0cm, width=1\linewidth]{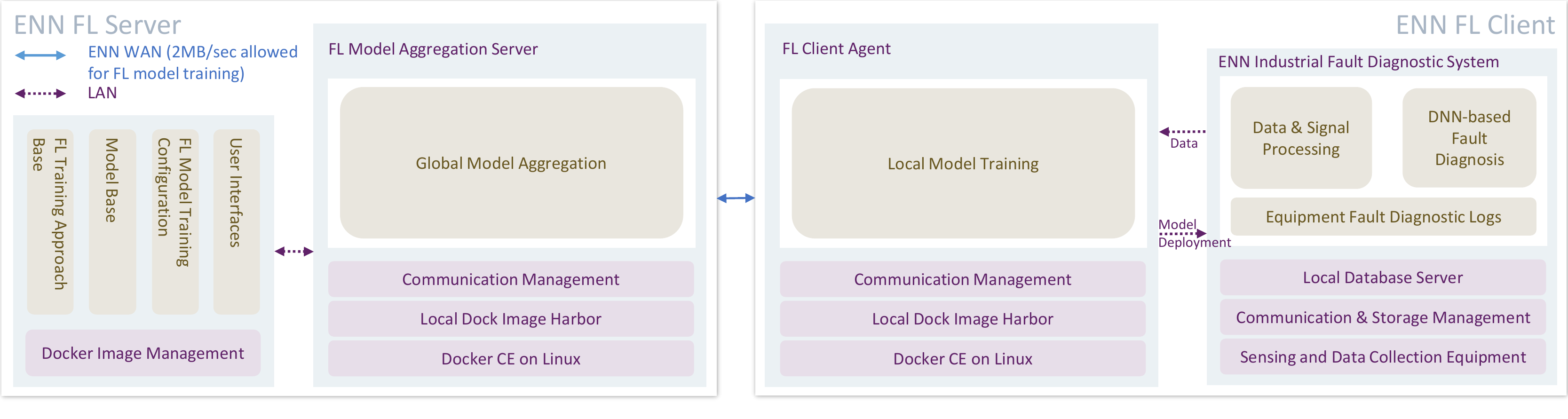}
  \caption{An overview of the ENN FL model training platform - Client-Server FL Subsystem.}\label{fig:ENN_FL}
\end{figure*}

\begin{figure}[t!]
  \centering
  \includegraphics[width=1\linewidth]{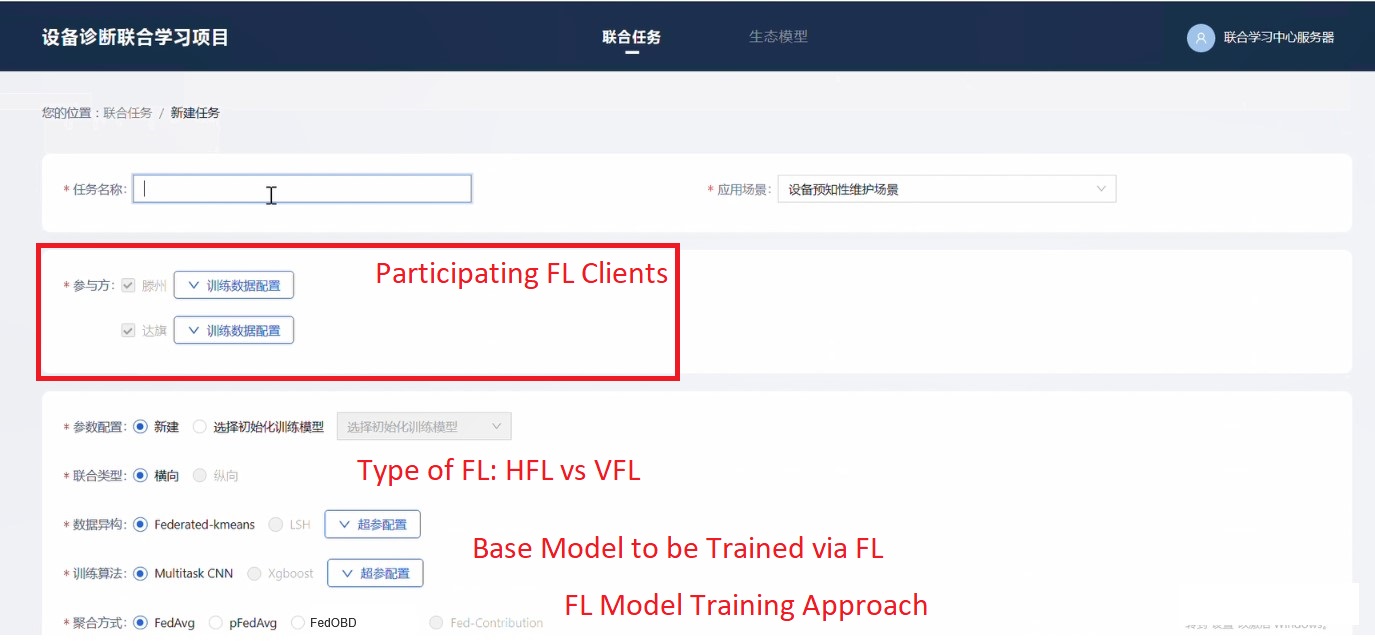}
  \caption{An example user interface of the ENN Client-Server FL Subsystem (configuring FL training).}\label{fig:UI1}
\end{figure}

\begin{figure}[t!]
  \centering
  \includegraphics[width=1\linewidth]{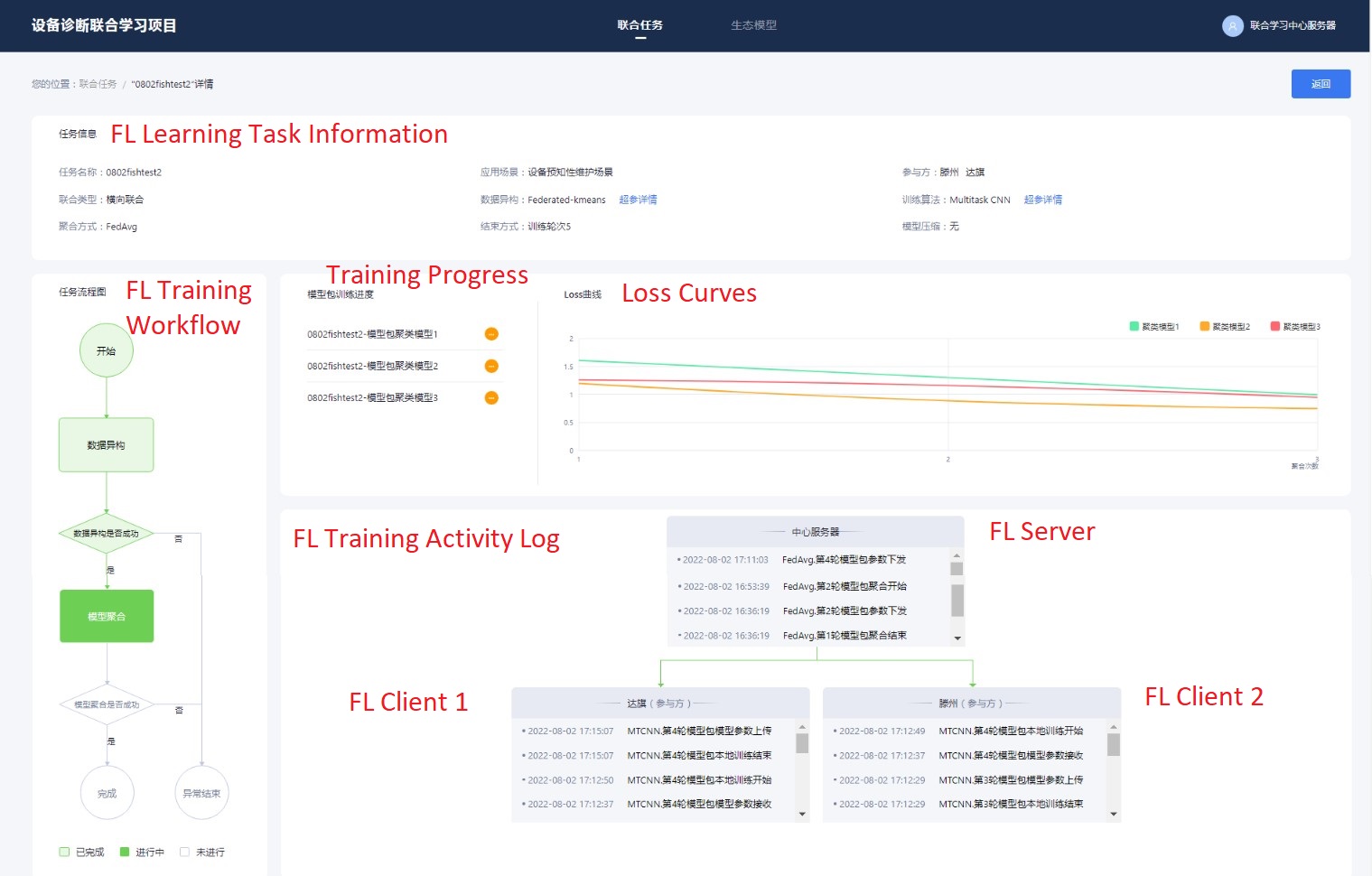}
  \caption{An example user interface of the ENN Client-Server FL Subsystem (FL training process visualization).}\label{fig:UI2}
\end{figure}

\begin{figure}[t!]
  \centering
  \includegraphics[width=1\linewidth]{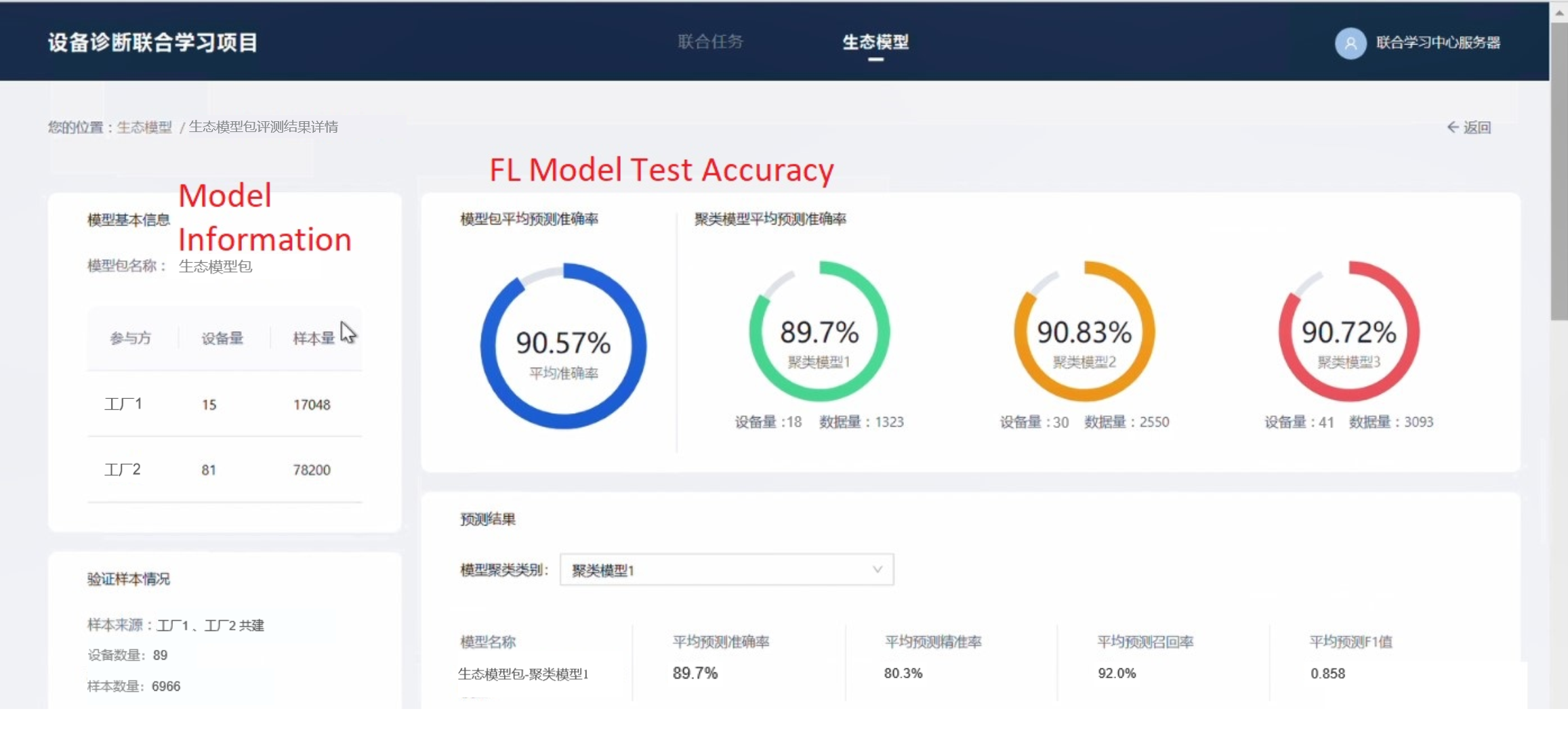}
  \caption{An example user interface of the ENN Client-Server FL Subsystem (model performance summary).}\label{fig:UI3}
\end{figure}

Founded in 1989, ENN Group's business encompasses a diverse range of segments within the natural gas and green energy industry including distribution, trade, transportation and storage, production, and intelligent engineering, with the aim of creating modern energy systems that improve people's quality of life. Leveraging its diverse industrial ecosystem, ENN has been building up an industrial digital intelligence platform in order to empower the stakeholders involved (including more than 25 million household customers and over 200,000 enterprise customers in 20 provinces across China). In this section, we provide detailed descriptions of the ENN FL model training platform, which is part of its industrial digital intelligence platform.

The overall flow of this platform is shown in Figure \ref{fig:Overview}. Enterprise customers under the ENN Group (e.g., coal chemical plants) deploy sensing devices within their factories to monitor equipment operation and collect data. These data are stored locally within the data silo. Standard data preprocessing (e.g., Fourier transform, wavelet transform) is carried out locally to prepare the data for analysis and model training. Nevertheless, as equipment faults do not occur frequently, such data tend to be sparse and biased within each data silo. The ENN Group offers its enterprise customers and subsidiaries from the same industry sectors the option to join FL to collaboratively train fault diagnostic models.
Our focus is on its FL model training subsystem with a client-server architecture. It consists of two components (Figure \ref{fig:ENN_FL}): 1) the ENN FL Server, and 2) the ENN FL Client.

\subsection{ENN FL Server}
The ENN FL Server hosts the FL model aggregation server and a set of utility modules supporting the operations which system administrators need to perform. It allows the system administrators to select from a range of FL model training and aggregation approaches (e.g., FedAvg \cite{pmlr-v54-mcmahan17a}, \methodname{}) incorporated into the system, select the base model to be trained via FL, and configure the FL model training process. These operations can be carried out through a dedicated set of user interfaces. Once the configuration steps are completed, the information is sent to the FL model aggregation server for execution via a local area network (LAN). As the LAN is a dedicated communication channel for FL related operations, it does not place any restrictions on bandwidth usage.

The FL model aggregation server is implemented following a modular design to enable it to host alternative FL model aggregation approaches. It takes local model updates from FL clients as inputs, and produces a global FL model as the output, while making decisions on whether additional rounds of FL training are required. Communication with the clients is managed by the Communication Management module, which can accommodate special requirements (e.g., the need for compressing the transmission via stochastic quantization \cite{alistarh2017qsgd}). This is because communications between the ENN FL Server and the ENN FL Clients (which are deployed in different factories) take place over the ENN wide area network (WAN), which is not dedicated to just FL model training. Transmissions for other operational and business purposes also go through the ENN WAN. Thus, this channel places a limit of up to 2 MB/sec to be used for FL model training purposes, which severely restricts the speed of training large-scale models through FL.

Figure \ref{fig:UI1} to Figure \ref{fig:UI3} illustrate the user interfaces (UIs) through which the ENN FL model training platform visualizes the FL training process for the system administrators. As it is designed for Chinese speaking users, we have annotated regions in the UIs to highlight key design features. Figure \ref{fig:UI1} shows the screen for the administrator to configure the FL training process for a particular model by specifying important parameters (e.g., mode of federated learning, the selected model training and aggregation approach).
Once FL training commences, the training activities are visualized in Figure \ref{fig:UI2} for the administrators to monitor. In the example in Figure \ref{fig:UI2}, one FL server and two FL clients are involved. Activities performed by the server and each client are listed in the corresponding box, making it easy to scroll back and forth for an overview, and drill down into each record for more detailed information. The overall training progress is illustrated in the FL training flowchat on the left hand side panel.
After training is concluded, a summary of the performance of the resulting FL model as shown in Figure \ref{fig:UI3} is presented to the administrators. Detailed FL model training activities and performance evaluation results are stored by the platform to support review and auditing in the future.

\subsection{ENN FL Client}
Typically, each ENN FL Client is deployed in an industrial facility (e.g., factory, power plant, coal chemical plant). In the application of our focus, industrial fault diagnostics, each facility deploys a set of sensors to monitor the industrial equipment and collect the necessary data. The Communication \& Storage Management module aggregates the data and stores them into the Local Database Server in the correct format. On top of this infrastructure, the ENN Industrial Fault Diagnostic System carries out data and signal processing, and performs fault prediction using a DNN-based model. In order to leverage industrial fault data collected by different facilities, this DNN-based fault prediction model is to be trained through FL.

Each ENN FL Client is incorporated with an FL Client Agent module. Similar to the ENN FL Server, it is also implemented following the modular design approach to enable it to host alternative FL model training and updating approaches. The dedicated Communication Management module is also included to accommodate special transmission requirements and to comply with the 2 MB/sec transmission bandwidth usage limit placed on the ENN WAN for FL model training.

\section{Use of AI Technology}
\begin{figure*}[t!]
  \centering
  \includegraphics[width=1\linewidth]{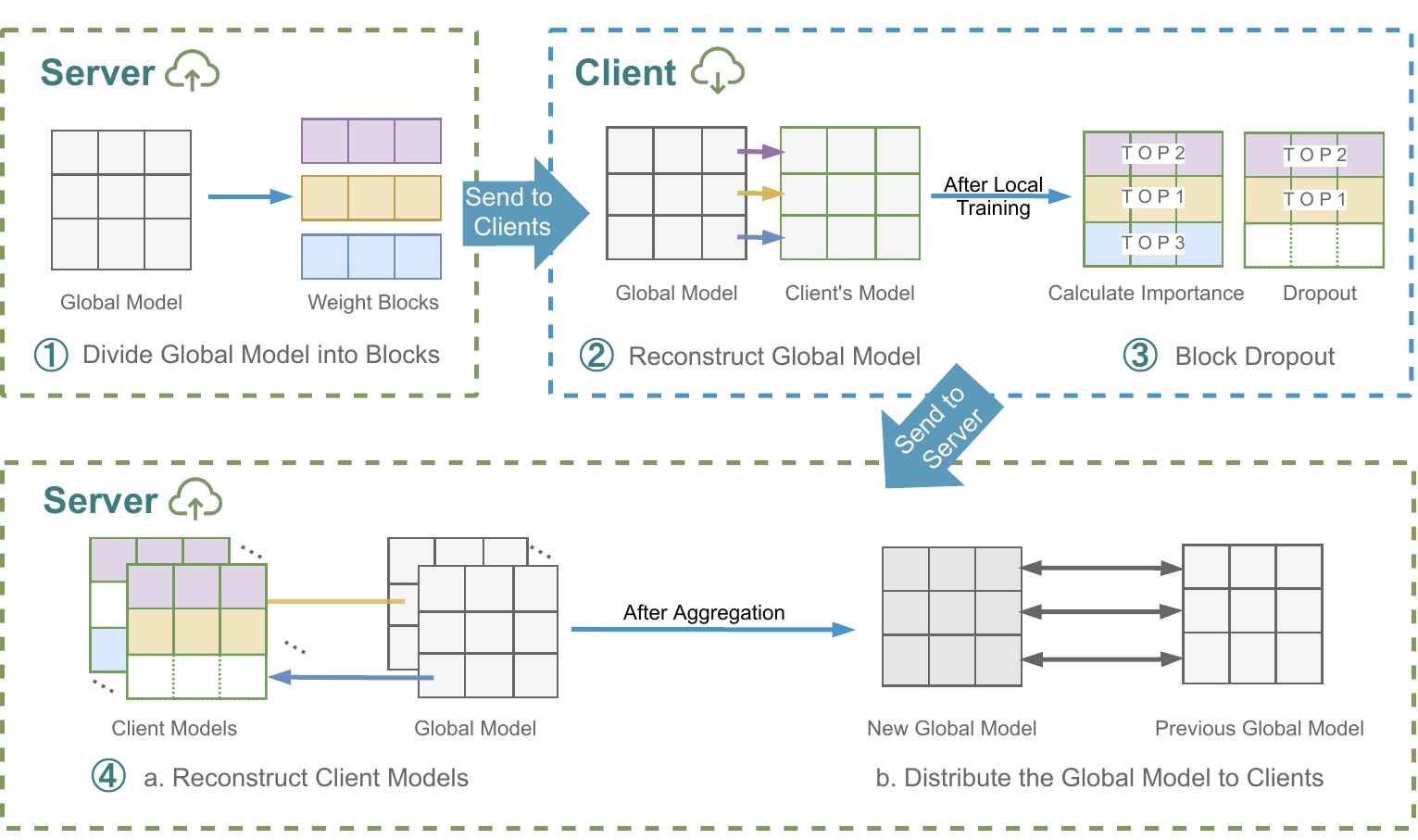}
  \caption{The workflow of the \methodname{} approach.}\label{fig:FedOBD}
\end{figure*}

In this section, we describe the AI Engine of the ENN FL model training subsystem, which is based on \methodname. The workflow of the AI Engine is illustrated in Figure \ref{fig:FedOBD}. \methodname{} \cite{FedOBD} enables the FL server and each FL client to determine the most important subset of parameters of a large-scale deep neural network (DNN) model (which are organized into semantic blocks) to be sent back and forth during FL model training, thereby reducing the communication overhead incurred. Currently, it only supports horizontal federated learning (HFL) in which data owners have large overlaps in the feature space, but little overlap in the sample space \cite{FL2019}.

In a client-server HFL system, there are in general $n$ clients who can participate in FL model training. Each client $i$ has a local dataset $D_i = \left \{  \left (  \vect{x}_j, \vect{y}_j \right )  \right \}_{j=1}^{M_i}$. $\vect{x}_j$ denotes the $j$-th local training sample. $\vect{y}_j$ denotes the corresponding ground truth label of $\vect{x}_j$. $M_i$ denotes the total number of data samples in $D_i$. 
The aim of HFL is to solve the following optimization problem under the aforementioned setting:
\begin{equation}
  \min_{\boldsymbol{w}\in \boldsymbol{W}}\sum_{i=1}^{n}\frac{M_i}{M} \mathcal{L}_{i}(\boldsymbol{w};D_i).
\end{equation}
Here, $\boldsymbol{W}$ denotes the parameter space determined by a given neural network. $M \coloneqq \Sigma_{i=1}^n M_i$ denotes the total number of samples. $\mathcal{L}_{i}(\boldsymbol{w};D_i) \coloneqq \frac{1}{M_i} \sum_{j=1}^{M_i} \ell(\boldsymbol{w};\vect{x}_j, \vect{y}_j)$ denotes the local loss of a given client $i$.

\subsection{Opportunistic Block Dropout (OBD)}
A DNN can be divided into semantic blocks consisting of consecutive layers. Under the \methodname{} FL model training approach, important semantic blocks in a DNN are identified at the end of any given round of FL client local training or FL server aggregation. Once this is done, semantic blocks are selected in descending order of their importance until a dropout rate pre-specified by the system administrators has been reached. Then, only these selected blocks of the DNN, instead of the entire model, are transmitted to facilitate FL model training. This design of \methodname{} is different from existing dropout-based FL training approaches including FedDropoutAvg approach~\cite{FedDropoutAvg} and Adaptive Federated Dropout (AFD) approach~\cite{fed-dropout}. Both of which randomly select individual model parameters to be dropped out without organizing the model into semantic blocks first, making the resulting model difficult to compress during transmission.

\methodname{} can support popular NN architectures when decomposing the models. Layer sequences such as $\langle$Convolution, Pooling, Normalization, Activation$\rangle$ are commonly found in convolutional neural networks (CNNs). Encoder layers are commonly found in Transformer based models. Other NNs can include basic building blocks which can be used to divide a given model into blocks. Other layers which cannot be grouped into commonly found functional block patterns can be treated as singleton blocks.
\methodname{} uses the Mean Block Difference (MBD) metric to measure block importance. It can be computed as follows:
\begin{equation}
  \funname{MBD}(\boldsymbol{b}_{r-1}, \boldsymbol{b}_{r,i}) \coloneqq \frac{\norm{\funname{vector}(\boldsymbol{b}_{r-1}) - \funname{vector}(\boldsymbol{b}_{r,i})}_2}{\funname{NumberOfParameters}(\boldsymbol{b}_{r-1})}. \label{eq:MBD}
\end{equation}
$\boldsymbol{b}_{r-1}$ denotes the blocks of a previous model (e.g., the received global FL model). $\boldsymbol{b}_{r,i}$ denotes the corresponding blocks in an updated model (e.g., the local model trained by a client $i$ in the current round). $\funname{vector}$ is an operator that concatenates parameters from different layers of a block (if there are multiple layers involved) into a single vector. The larger the MBD value of a block, the more important the newer version of this block is.

\begin{algorithm}[t!]
  \caption{OBD}
  \label{algo:block}
  \SetKwInOut{KwIn}{Input}
  \SetKwInOut{KwOut}{Output}
  \SetKw{Continue}{continue}
  \KwIn{global model $\boldsymbol{w}_{r-1}$, local model $\boldsymbol{w}_{r,i}$ in client $i$, the set of identified block structures $\boldsymbol{B}$, dropout rate $\lambda\in [0,1]$. }
  \KwOut{retained blocks.}

  $\textnormal{important\_blocks} \leftarrow\funname{MaxHeap}()$;

  \ForEach{$\boldsymbol{b} \in \boldsymbol{B}$}
  {
    $\textnormal{important\_blocks}[\funname{MBD}(\boldsymbol{b}_{r-1}, \boldsymbol{b}_{r,i})]       \leftarrow  \boldsymbol{b}_{r,i} $;
  }
  $\textnormal{revised\_model\_size} \leftarrow 0$; \\
  $\textnormal{retained\_blocks} \leftarrow \funname{List}()$; \\
  \While{$\textnormal{important\_blocks}$}
  {
    $\boldsymbol{b}_{r,i} \leftarrow \textnormal{important\_blocks}.\funname{pop}()$;  \\
    $\textnormal{new\_size} \leftarrow \textnormal{revised\_model\_size} + \lonenorm{\funname{vector}(\boldsymbol{b}_{r,i})}$; \\
    \If{ $\textnormal{new\_size} > (1-\lambda)\lonenorm{\funname{vector}(\boldsymbol{w}_{r,i})}$ }
    {
      \Continue{}; \\
    }
    $\textnormal{revised\_model\_size} \leftarrow  \textnormal{new\_size}$; \\
    $\textnormal{retained\_blocks}.\funname{append}(\boldsymbol{b}_{r,i})$;
  }
  \Return{$\textnormal{retained\_blocks}$};
\end{algorithm}
The Opportunistic Block Dropout (OBD) algorithm of \methodname{} is shown in Algorithm~\ref{algo:block}. Before a model is sent out either by the FL server or client, each block is assigned an importance score. To achieve this goal, the sending entity stores $\boldsymbol{w}_{r-1}$ which can be used to compare with the current model $\boldsymbol{w}_{r,i}$ in a block by block fashion by following Eq. \eqref{eq:MBD}.

Once the MBD values for all the blocks have been computed,~\methodname{} determines which blocks to retain and which to be dropped (Lines 5-15) by ranking in descending order of their MBD values (with the \funname{MaxHeap} data structure), and putting the blocks into the retained\_blocks list one by one until the size of the revised model reaches $(1-\lambda)\lonenorm{\funname{vector}(\boldsymbol{w}_{r,i})}$. $\lambda\in[0,1]$ is the dropout rate (where 1 indicates the entire model is dropped out, and 0 indicates no dropout). The retained blocks are then quantized in preparation for transmission. Each retained block is stored in the form of the differences between the corresponding parameter values in block $\boldsymbol{b}_{r}$ and block $\boldsymbol{b}_{r-1}$.

\subsection{Overall Workflow of \methodname}
The overall workflow of the \methodname{} approach is illustrated in Figure \ref{fig:FedOBD}.
\begin{enumerate}
  \item \textbf{Model Distribution}: If it is the first time the initialized global FL model is distributed to the FL clients, the entire model is sent out by the FL server. Otherwise, the server performs OBD to determine the list of important blocks of the model to be retained based on the given dropout rate, and only sends out the quantized version of these retained blocks to the clients.
  \item \textbf{Reconstruction of the Global Model}: When the retained blocks from the server in round $r$ is received by a client, it combines them with unchanged (i.e., not transmitted) blocks from the global model $\boldsymbol{w}_{r-1}$ which it has received in the previous round to reconstruct $\boldsymbol{w}_{r}$. Local training is then carried out based on this reconstructed global model.

        \begin{figure*}[t!]
          \centering
          \includegraphics[clip, trim=0cm 10cm 0cm 0cm, width=1\linewidth]{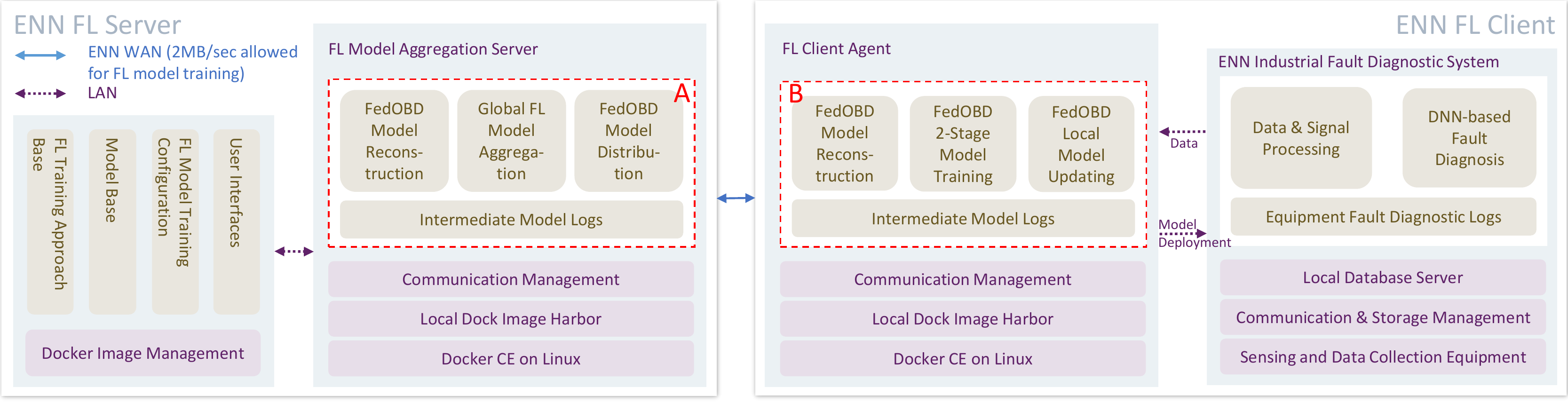}
          \caption{Deployment of \methodname{} into the ENN Client-Server FL Subsystem.}\label{fig:ENN_FL_FedOBD}
        \end{figure*}

  \item \textbf{2-Stage Local Model Training}:~\methodname{} is implemented as a two-stage training process. In the first stage, small local epochs are used. OBD selects important blocks to be transmitted, and these blocks are further compressed via quantization. In this way,~\methodname{} encourages frequent aggregation to prevent overfitting in a communication efficient manner. As the global FL model approaches convergence,~\methodname{} transits into the second stage in which FL model training is switched to the training-aggregation mode. It consists of a single round of training with more local epochs with FL model aggregation being executed at the end of each epoch. In this way, the global FL model is fine-tuned through more frequent aggregations in order to improve its performance.
  \item \textbf{Model Aggregation}: When the retained blocks from a client $i$ in round $r$ is received, the FL server combines them with unchanged (i.e., not transmitted) blocks from the previous global model $\boldsymbol{w}_{r-1}$ to reconstruct $\boldsymbol{w}_{r,i}$. The reconstructed local model updates for all clients are then aggregated into a new global FL model $\boldsymbol{w}_r$ following FedAvg~\cite{pmlr-v54-mcmahan17a}.
\end{enumerate}
Steps 1 to 4 are repeated until model convergence. For more details about \methodname{}, please refer to \cite{FedOBD}.

\section{Application Development and Deployment}
The \methodname{} framework has been developed based on the PyTorch~\cite{paszke2017automatic} framework by teams from the Trustworthy Federated Ubiquitous Learning (TrustFUL) Lab, Nanyang Technological University (NTU), Singapore, and the Digital Research Institute, ENN Group, Beijing, China.
Before deploying the AI Engine, we have evaluated it against the classic FedAvg and four state-of-the-art efficient FL model training approaches. They are:
\begin{enumerate}
  \item \textbf{FedAvg}~\cite{pmlr-v54-mcmahan17a}: It is a classic FL approach which does not involve any compression or dropout operation.
  \item \textbf{SignSGD}~\cite{bernstein2018signsgd}: It is a distributed gradient compression approach which only requires the signs of the gradients of client model updates to be sent to the FL server. The server aggregates the gradients by majority voting.
  \item \textbf{FedPAQ}~\cite{reisizadeh2020fedpaq}: It is a stochastic quantization-based FL approach designed to reduce communication overhead.
  \item \textbf{Adaptive Federated Dropout (AFD)}~\cite{fed-dropout}: It is an FL approach that optimizes server-client communications and computation costs jointly. Each client trains a selected subset of the global model parameters. We adopt the Single-Model Adaptive Federated Dropout (SMAFD) variant for comparison.
  \item \textbf{FedDropoutAvg}~\cite{FedDropoutAvg}: It is an FL approach that randomly drops out a subset of model parameters as well as randomly drops out some clients before performing FedAvg.
\end{enumerate}

To compare the efficiency and performance of these approaches under different FL settings, we designed FL scenarios involving 10 clients to perform image classification on the CIFAR-10 datasets~\cite{krizhevsky2009learning} and sentiment classification on the IMDB dataset~\cite{maas2011learning}.
The original test data are split uniformly to form separate validation and test datasets. The local training and validation dataset of each client are sampled following an i.i.d. setting. The FL server holds the test dataset.
For the image classification tasks under CIFAR-10, we use FL to train a DenseNet-40~\cite{huang2017densely} base model which contains around 190,000 model parameters. For the sentiment classification task under IMDB, we use FL to train a Transformer based classification model consisting of 2 encoder layers followed by a linear layer with around 17 million model parameters.

\begin{table}[b!]
  \resizebox*{1\columnwidth}{!}{
    \begin{tabular}{|l|c|c|c|c|}
      \hline
                    & \multicolumn{2}{c|}{CIFAR-10} & \multicolumn{2}{c|}{IMDB}                                         \\ \cline{2-5}
                    & Communication                 & Test                      & Communication      & Test             \\
                    & Overhead (MB)                 & Accuracy                  & Overhead (MB)      & Accuracy         \\ \hline
      FedAvg        & 1,467.30                      & 89.36\%                   & 131,494.20         & 84.68\%          \\ \hline
      SignSGD       & 7,208.75                      & 60.18\%                   & 327,100.00         & 50.21\%          \\ \hline
      FedPAQ        & 463.54                        & 89.22\%                   & 41,541.05          & 82.94\%          \\ \hline
      SMAFD         & 522.23                        & 17.38\%                   & 63,365.16          & 63.94\%          \\ \hline
      FedDropoutAvg & 512.20                        & 87.53\%                   & 43,992.33          & 84.17\%          \\ \hline
      \methodname{} & \textbf{101.59}               & \textbf{90.17\%}          & \textbf{12,899.03} & \textbf{84.98\%} \\ \hline
    \end{tabular}
  }
  \caption{Pre-deployment experiment results.}\label{tb:results}
\end{table}

The results are summarized in Table \ref{tb:results}. Communication overhead is computed as the product between the average fraction of model size (in MBs) transmitted per FL training step and the total number of FL training steps required. It can be observed that \methodname{} significantly outperforms all existing approaches in terms of reducing communication overhead. In terms of test accuracy of the resulting model, \methodname{} significantly outperforms existing compression or dropout-based efficient FL training approaches, achieving comparable performance to FedAvg which does not engage in any model compression or dropout. The results helped the design team make the decision to adopt \methodname{} as the FL approach for training large-scale industrial fault diagnostic models in the ENN platform.

\begin{figure*}[t!]
  \centering
  \includegraphics[clip, trim=0cm 4cm 5.4cm 0cm, width=1\linewidth]{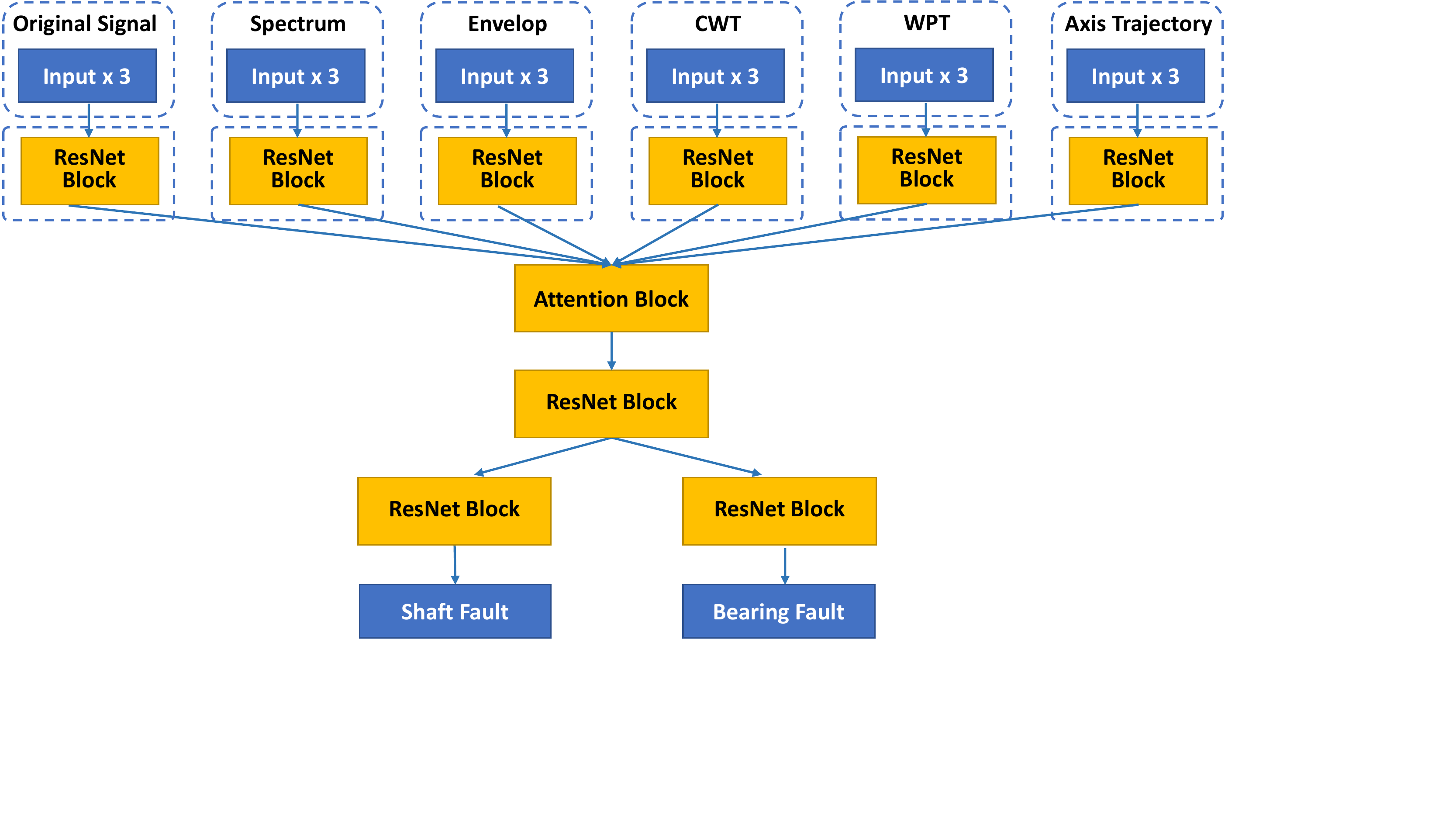}
  \caption{The architecture of the fault prediction deep neural network model.} \label{fig:DNN}
\end{figure*}

\methodname{} is deployed into the ENN Client-Server FL Subsystem as an alternative FL training approach the system administrators can choose to use. It is added into the user interface as a new option that can be selected during FL training configuration (Figure \ref{fig:UI1}). Once \methodname{} is selected as the FL training approach from the user interface, the model reconstruction method, the global model aggregation method and the the block dropout and model distribution method under \methodname{} are loaded into the FL server. The intermediate model logs module is also included in the FL server to store the global model from the previous round to facilitate the evaluation of block importance by \methodname{}, as illustrated in Figure \ref{fig:ENN_FL_FedOBD} (within the dashed rectangle A). At the same time, the model reconstruction method, the 2-stage model training method and the the block dropout and model uploading method under \methodname{} are loaded into the FL
Client Agent. Similarly, the intermediate model logs module is included in the FL Client Agent to store the global model from the previous round to facilitate the evaluation of block importance by \methodname{}, as illustrated in Figure \ref{fig:ENN_FL_FedOBD} (within the dashed rectangle B). In this way, \methodname{} is incorporated into the ENN FL platform.

\section{Application Use and Payoff}
\methodname{} has been deployed in ENN Group since February 2022 as part of its intelligent industrial fault
diagnostics platform. Since its deployment, the company has switched from the original FedAvg based FL model training and aggregation approach to \methodname{} for part of its enterprise customers. So far,~\methodname{} has been used to help two well-established coal chemical plants located in two cities in China to train AI models for industrial fault prediction.\footnote{At the request of our industry partners, the identities of these factories are withheld.}

Figure \ref{fig:DNN} illustrates the architecture of the DNN adopted by the ENN Group for fault prediction in the current deployment cycle. The model consists of a variety of vibration signal analysis methods to preprocess the original vibration signals to improve the interpretability of the model. Among them, the spectrum and the axis trajectory can show the frequency doubling component related to the rotating speed in the signal, which is helpful for diagnosing shaft faults. Envelope, continuous wavelet transform (CWT) and wavelet packet transform (WPT) can extract high-frequency impact components from the signal, which is helpful for diagnosing bearing and gearbox related faults. The attention block is used to adjust the importance of the features from different input blocks. At the end, two ResNet blocks are used to diagnose the two major types of faults, respectively. The model contains 29 million model parameters.


~\methodname{} is configured with the same hyperparameters as the previous FedAvg FL model training approach in the system. Specifically, a local learning rate of $10^{-5}$ is used in the ENN FL Clients. A quantization weight of $0.01$ is used for quantizing uploaded and distributed models. Only a single local epoch is used in FedAvg and the first stage of~\methodname{}. For the second training stage of~\methodname{}, two local epochs are used. 

\begin{table}[t!]
  \centering
  \resizebox*{1\columnwidth}{!}{
    \begin{tabular}{|l|c|c|}
      \hline
      ENN FL Training         & Communication & Test               \\
      Approach                & Overhead (MB) & F1 Score           \\\hline
      Previous AI Engine      & $368,407.60$  & $85.52 \pm 0.83\%$ \\
      \hline
      \methodname{} AI Engine & $104,188.50$  & $85.02 \pm 0.23\%$ \\
      \hline
    \end{tabular}
  }
  \caption{Deployment results.} \label{tb:results2}
\end{table}

Table \ref{tb:results2} shows the average communication overhead of training the fault prediction model until convergence for each round of model update, as well as the average test F1 score of the resulting models thus far into the deployment period under \methodname. The results under the Previous AI Engine reflects the same items but were collected during system operations in 2021 prior to switching to \methodname. It can be observed that \methodname{} saves communication overhead by $71.72\%$, while achieving comparable model performance. Due to the 2 MB/sec bandwidth usage limit placed on FL model training related communications by ENN Group, under the previous AI Engine, it took more than 2 days (around 52 hours) to train an updated version of the fault prediction model involving the two factories. Under the \methodname{} AI Engine, this time is reduced to about half a day (around 14.5 hours).


As new data generated by the equipment monitoring sensing devices are continually accumulated over time from and new equipment can be deployed in the factories from time to time, it is necessary to frequently retrain the fault prediction model via FL to keep it update to date. The deployment of \methodname{} has cut down the model training time by more than half while maintaining model performance, thereby enabling timely retraining of the model for enhanced safety and efficiency of operation.

\section{Maintenance}
The AI Engine follows a modular design approach to achieve separation of concerns.
Thus far into the deployment period, although there have been changes in personnel access rights and operating parameters in the system as well as frequent retraining of the fault prediction model via \methodname{}, such changes have not necessitated any AI maintenance task for the AI Engine.

\section{Lessons Learned During Deployment}
During the process of deploying the \methodname{} approach, there are several lessons worth sharing.

Firstly, the quality of training data is important to training effective industrial fault prediction models. The data cleaning and preprocessing steps by each participating data owner play an important role in model training. As data preprocessing is still mainly performed based on human experience, it can be expensive and prone to human errors. Since not all industry data owners have an in-house data science team, it could be a challenge to obtain preprocessed local data with consistently high quality.
Thus, it could be useful to adopt privacy-preserving data selection approaches such as \cite{Li-et-al:2021} to automate this process.

Secondly, as industrial fault prediction is an important application with high impact on operation safety and continuity, industry partners prefer some degrees of model interpretability. Although there exist parameter dropout-based FL approaches that can improve training efficiency \cite{fed-dropout,FedDropoutAvg}, their lack of interpretability on parameter dropout decisions hinders industry adoption. The block importance values produced during the intermediate steps of \methodname{} are helpful in providing the decision-makers with much needed transparency to alleviate such concerns.

Last but not least, as time goes by, new monitoring data from the participating factories will continue to be generated. Previously unencountered fault types may emerge. In addition, the data distribution might also change, resulting in concept drift \cite{Lu-et-al:2018}. These factors can be especially pronounced when new machines are incorporated into the factories. Thus, the performance of previously trained FL models can deteriorate in the face of these factors. Therefore, appropriate incremental training and updating strategies of the FL model need to be put in place to ensure successful deployment.


\section{Conclusions and Future Work}
In this paper, we reported on our experience using a dropout-based technique to enhance efficient collaborative training of large-scale deep models through federated learning for industrial fault diagnostic models involving multiple factories. We developed the \methodname{} FL model training and aggregation approach, which leverages a novel opportunistic importance-based semantic block dropout method in combination with quantization-based FL model parameter compression to drastically reduce communication overhead while preserving model performance. Since its deployment in February 2022 in ENN Group, \methodname{} has helped two well-established coal chemical plants in two cities in China to train machine learning models for fault diagnostics in order to support predictive maintenance, and has made significant positive impact on ENN Group's operations.

In future, we will enhance the robustness of \methodname{} against malicious FL participants \cite{Lyu-et-al:2020survey}. We will also explore how to link block importance evaluation with FL client contribution evaluation to enhance fairness \cite{Shi-et-al:2021}, as well as personalizing the resulting models \cite{Tan-et-al:2022TNNLS} to FL participants.
Eventually, we aim to incorporate \methodname{} into an opensource FL framework such as Federated AI Technology Enabler (FATE) \cite{FATE} and make it available to more developers, researchers and practitioners.

\section{Acknowledgements}
This research is supported, in part, by the National Research Foundation, Singapore under its AI Singapore Programme (AISG Award No: AISG2-RP-2020-019); the RIE 2020 Advanced Manufacturing and Engineering (AME) Programmatic Fund (No. A20G8b0102), Singapore; Nanyang Assistant Professorship (NAP); and Future Communications Research \& Development Programme (FCP-NTU-RG-2021-014). Any opinions, findings and conclusions or recommendations expressed in this material are those of the author(s) and do not reflect the views of National Research Foundation, Singapore.

\bibliography{aaai23}

\end{document}